\documentclass[letterpaper, 10 pt, conference]{ieeeconf}  % Comment this line out if you need a4paper

\usepackage{graphics} % for pdf, bitmapped graphics files
\usepackage{epsfig} % for postscript graphics files
\usepackage{mathptmx} % assumes new font selection scheme installed
\usepackage{times} % assumes new font selection scheme installed
\usepackage{amsmath} % assumes amsmath package installed
\usepackage{amssymb}  % assumes amsmath package installed
\usepackage{hyperref}
\usepackage[nospace]{cite}
\usepackage{multirow}
\usepackage{booktabs}
\usepackage{amssymb}
\usepackage{graphicx}
\usepackage{footnote}
\usepackage{newtxmath}
\makesavenoteenv{tabular}
\usepackage[labelformat=simple]{subcaption}
\makeatletter
\def\thickhline{%
  \noalign{\ifnum0=`}\fi\hrule \@height \thickarrayrulewidth \futurelet
   \reserved@a\@xthickhline}
\def\@xthickhline{\ifx\reserved@a\thickhline
               \vskip\doublerulesep
               \vskip-\thickarrayrulewidth
             \fi
      \ifnum0=`{\fi}}
\makeatother

\IEEEoverridecommandlockouts                              % This command is only needed if 
\newlength{\thickarrayrulewidth}
\title{\LARGE \bf
See Yourself in Others:\\ Attending Multiple Tasks for Own Failure Detection
}

\author{Boyang Sun$^{*}$, Jiaxu Xing$^{*}$, Hermann Blum, Roland Siegwart, and Cesar Cadena
\thanks{* Authors contributed equally to this work. This work was partially supported by the Hilti Group and the National Center of Competence in Research (NCCR) Robotics through the Swiss National Science Foundation.}
\thanks{All the authors are with the Autonomous Systems Lab, ETH Zurich, 8092, Switzerland. {\tt\{boysun, jixing, blumh, rsiegwart, cesarc\}@ethz.ch}
}
}

\begin{document}

\maketitle
\thispagestyle{empty}
\pagestyle{empty}

%%%%%%%%%%%%%%%%%%%%%%%%%%%%%%%%%%%%%%%%%%%%%%%%%%%%%%%%%%%%%%%%%%%%%%%%%%%%%%%%
\begin{abstract}

Autonomous robots deal with unexpected scenarios in real environments. Given input images, various visual perception tasks can be performed, e.g., semantic segmentation, depth estimation and normal estimation. These different tasks provide rich information for the whole robotic perception system. All tasks have their own characteristics while sharing some latent correlations. However, some of the task predictions may suffer from the unreliability dealing with complex scenes and anomalies.
We propose an attention-based failure detection approach by exploiting the correlations among multiple tasks. The proposed framework infers task failures by evaluating the individual prediction, across multiple visual perception tasks for different regions in an image. The formulation of the evaluations is based on an attention network supervised by multi-task uncertainty estimation and their corresponding prediction errors. Our proposed framework\footnote{Code link \url{https://github.com/ethz-asl/uncertainty_with_multiple_tasks}.} 
generates more accurate estimations of the prediction error for the different task's predictions.

\end{abstract}

%%%%%%%%%%%%%%%%%%%%%%%%%%%%%%%%%%%%%%%%%%%%%%%%%%%%%%%%%%%%%%%%%%%%%%%%%%%%%%%%
\section{INTRODUCTION}

% Introduce the background of the paper: for robotics perception (good performance on cases with similar distribution to the training set) , dealing with open and unconstrained world is still an open problem
Extensive research has shown that visual information is an important component in autonomous driving and many robotic perception systems\cite{Grigorescu_2020, cesar16, Garg_2020}.
Autonomous agents utilize the information from various learning-based visual perception predictions. Existing works have shown good performance on cases where the deployment environment has similar distribution to the training set \cite{bulusu2020anomalous}.
However, many state-of-the-art deep learning approaches still face the lack of ability in dealing with open and unconstrained world \cite{godard2017unsupervised, godard2019digging, zhu2019improving}, and will produce failures especially in unseen environments \cite{Gal2016Uncertainty}. Thus, a method to detect prediction failures of various robotics visual perception tasks is crucial for safe robotic deployments. With higher introspection capabilities, autonomous robots will be more controllable in safety-critical scenarios.

\begin{figure}
    \centering
    \includegraphics[scale=0.275]{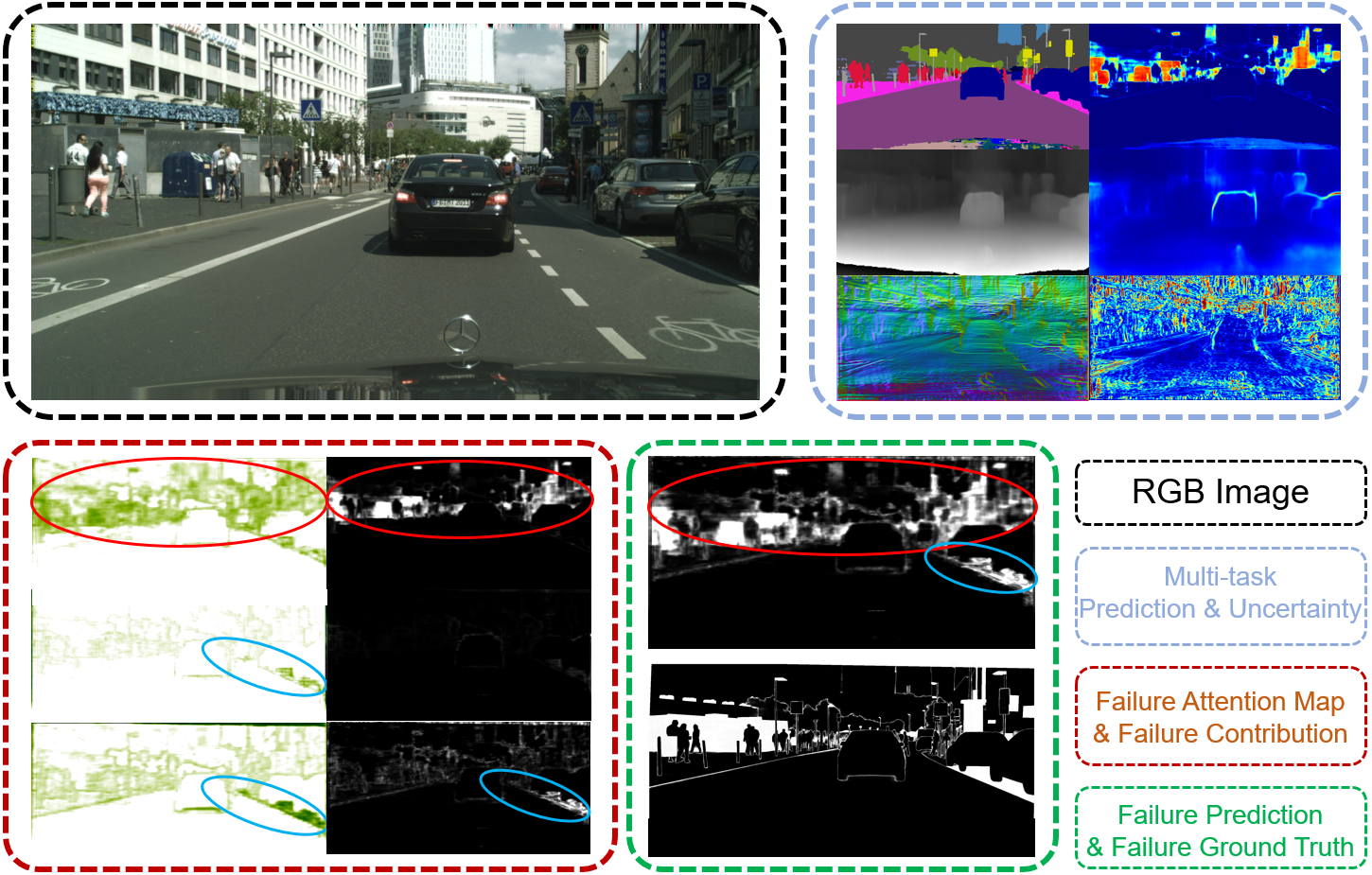}
    \caption{\textbf{Example Result of Our Approach.} Our method captures the high prediction uncertainty regions of a single task using multiple visual tasks. The result maintains the useful uncertainty estimation from the original task (highlighted areas in red circle). Moreover, beneficial from the multi-task setup, our approach captures the relevant information from other tasks (highlighted areas in blue circle) to compensate the missed failure regions.} 
    \label{fig:eval_1}
    \vspace{-5mm}
\end{figure}

% Specify the main content of this paper's work: we use multi-task, we use uncertainty, and we use attention mechanism - Exploit the complementary information from multiple tasks to improve the introspection capability (failure prediction) of the perception system on each single task.
This work focus on identifying failure predictions of various robotics perception tasks by exploiting the latent correlations among them. Those correlations have been recently used to improve tasks performance \cite{zamir2018taskonomy, Zamir2020CVPR}. %Based on such inspiration, we build a failure prediction algorithm using information across multiple perception tasks. For each task, we also consider its uncertainty of prediction as a reference of failure detection, and use these uncertainties to help with building an attention network.
Our basic idea is to exploit the complementary information from multiple tasks to improve the introspection capability of the perception system on every single task based on an attention mechanism. 
Our failure detection model has a \textit{unified} structure that attends the encoded multi-task feature maps with the \textit{expressive power} to perform failure prediction for different tasks. 

%TODO: compared with the uncertainty method we used.
%% Q1: Should we briefly mention how well our model performs compared to other methods (probably not SOTA)? Yes. We can see later if fits better here or in the RW

% Introduce the lack of single task & various of uncertainties: For single task, fail when the prediction is incorrect, for uncertainties, can not fit into various scenarios(robustness). Then briefly introduce our idea. This should follow the previous paragraph for the idea
% New version: Introducing the main component of our model: uncertainty, attention and multi-task, and why use them
Existing research recognizes uncertainty as a common measurement of the multi-task prediction’s confidence \cite{Gal2016Uncertainty}. The uncertainties are ideally correlated to the corresponding task prediction errors, which measures the reliability of the predictions. General uncertainty estimation methods are based on a single task, e.g., softmax entropy from semantic segmentation\cite{hendrycks2016baseline}. However, the quality of the uncertainty estimation is limited by several factors, such as environmental conditions (e.g., clean/foggy weather), anomalies, and more \cite{di2021pixel}. In addition, to avoid the limitation brought by single task uncertainty estimation, our model takes various tasks encoded predictions as input and computes an attention map for each single task. The attention values are modeled as the regional contribution of every task uncertainty to one certain task prediction error.
%This model works with various kinds of different distinct but correlated visual task predictions, such as semantic segmentation , depth estimation  normal estimation , instance segmentation , which we choose to use in this paper.

% Introduce how we set up experiments, how we evaluate, which metrics we choose, which dataset we use, and briefly summarize the result. And build a paper map.
To investigate the model robustness, we train our model on the Cityscapes \cite{Cordts2016Cityscapes} dataset and test it on several datasets with different distribution characteristics\cite{darkzurich, Zendel_2018_ECCV}. We evaluate the model for different tasks and compare it against several existing methods. We show that our approach outperforms all other failure prediction approaches. Moreover, our framework is flexible to the number and types of tasks with different task prediction $\&$ uncertainty estimation methods. A result example of our approach is shown in Figure \ref{fig:eval_1}.% We also demonstrate that, even with less input prediction entries, upon an acceptable drop of the performance, the model could still assign meaningful attention and result a reasonable error prediction. 
%not sure if we keep the contribution items here 

\bigskip
\bigskip

In summary, the contributions of this work are:
\begin{itemize}
    \item The first work to exploit the multiple visual tasks setup for detecting failures in their prediction in deployment. 
    \item A novel framework with an attention mechanism over the multiple visual tasks being deployed to extract the complementary information in their uncertainty estimates for failure detection.
    \item A thorough evaluation of design components and their influence in open world scenarios.

\end{itemize}

% \begin{figure}
%     \centering
%     \includegraphics[scale=0.3]{figures/pipeline.png}
%     \caption{\textbf{Result Overview} All major results of our approach are visualised in this figure. On the left side are the input-level entries, which contains original images, different task predictions and their prediction uncertainties. On the right side are the output-level results, which contains the attention maps and the task prediction error approximation. Details about each entry will be further explained in the report}
%     \label{fig:intro}
% \end{figure}
\section{Related Work}
\label{sec:citations}

% Overview of the related work: the top idea is to do failure detection(FD). We classified them into several sub-class: FD using basic uncertainty estimation methods(for semantic, depth, or in general, etc.); FD using learning based approach in recent works. Another sub-class is multi-task failure correction/performance improvement, by introducing several multi-task related paper, we say we borrow the idea of multi-task correlation, and use it as a basic idea for our failure detection.  

Works on single task failure prediction can be classified in failure detection from the estimated uncertainty and learning-based failure detection. Other works have exploited the multi-task setup to achieve better per-task prediction accuracy. However, to the best of our knowledge, failure prediction from multiple tasks has not been proposed prior to this work. 

\subsection*{Failure Detection via Uncertainty Estimation}
Uncertainty estimation has a close relation with failure detection and introspection. The uncertainty of the prediction results reflects the level of its confidence. And the intuition follows that a low-confidence prediction is likely a failure. Therefore, uncertainty estimation could be regarded as a reference to failures prediction. A conventional way to calculate the uncertainty is to directly analyze the distribution of the model prediction such as the \textit{softmax entropy}  \cite{lakshminarayanan2016simple}, or \textit{softmax distance} \cite{rottmann2019prediction} used in classification models.
Besides, \textit{image flipping} \cite{godard2017unsupervised} investigates the model results' difference in dealing with the original and flipped image. Finally, \textit{Bayesian estimation}  \cite{kendall2017uncertainties,ilg2018uncertainty,neal2012bayesian} estimates the uncertainty by sampling multiple models, e.g.  \textit{Monte-Carlo dropout} \cite{Gal2016Uncertainty, Poggi_CVPR_2020} captures the uncertainty by randomly dropping the connection between different layers.

\subsection*{Failure Detection via Learning-based methods}
Given the rapid development of deep learning. Neural network has become a possible option for failure detection task. Most of the failure detection methods in the visual perception area focus on detect semantic segmentation miss-classification. These methods can be roughly divided into two categories. One group directly trains detectors with failure cases\cite{hendrycks2018deep, kuhn2021pixel, marufur2021fsnet}. The other group uses re-synthesis methods \cite{lis2019detecting,haldimann2019not, di2021pixel, kuhn2021pixel} that rebuild image from semantic prediction, and capture the anomalies by comparing the rebuilt image and the original one.  

\begin{figure}[t]
    \centering
    \includegraphics[scale=0.27]{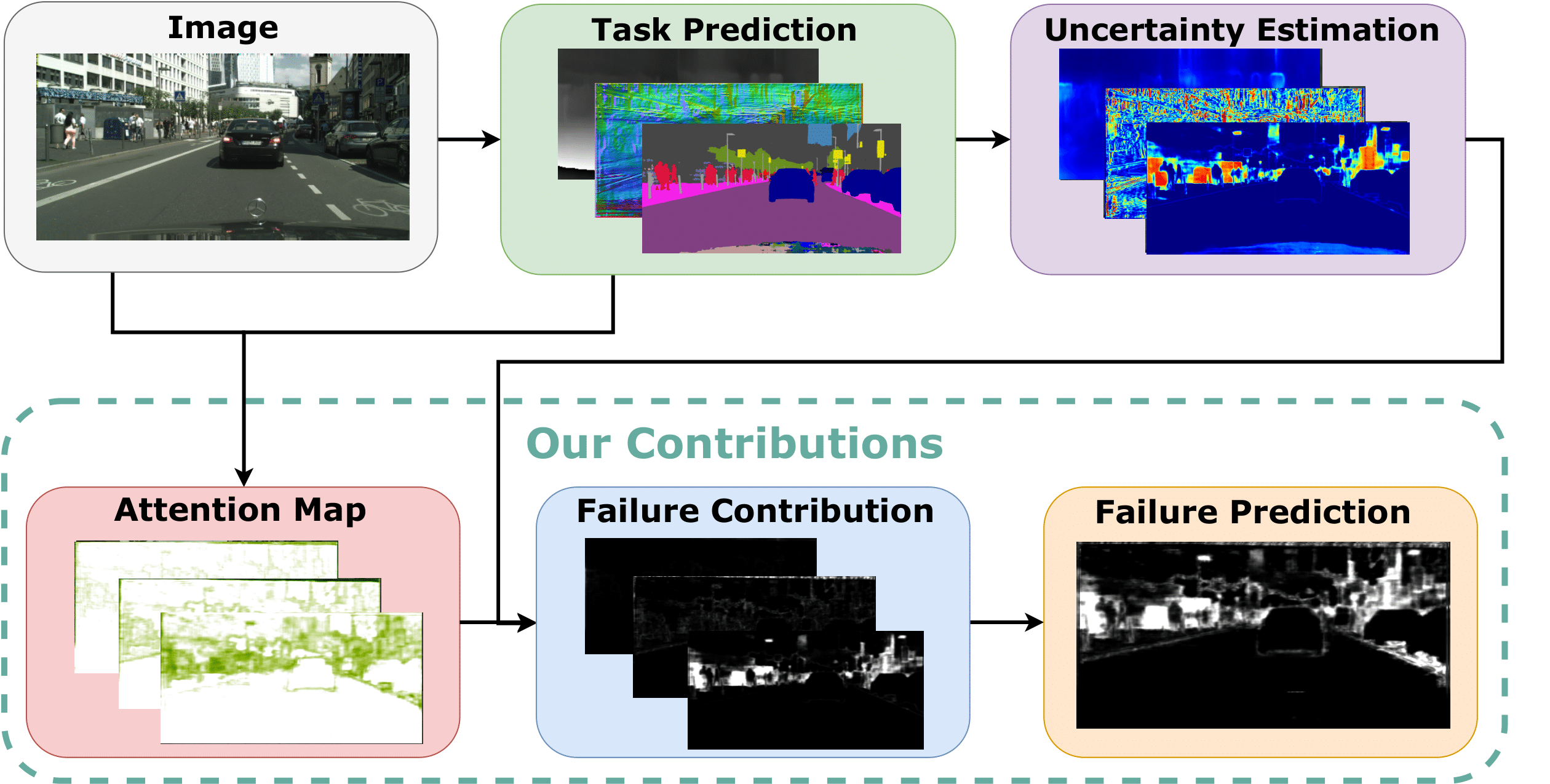}
    \caption{\textbf{Visual Tasks Failure Detection Framework.} Given an image and its multiple tasks predictions, our approach computes the attention maps to weigh the multiple tasks uncertainty estimations. This weighted sum of attention and uncertainty maps is our failure prediction for a chosen task.}
    \label{fig:overall}
    \vspace{-6mm}
\end{figure}

\subsection*{Learning from Multiple Tasks}
Prior works have already acknowledged the relation among different perception tasks \cite{relatedness, ubernet}. This latent correlated structure among visual tasks has been exposed in the work of \textit{Taskonomy} \cite{zamir2018taskonomy}. The utilization of cross-task relations also lies in the area of domain adaptation \cite{aytar2011tabula,patel2015visual}, transfer learning \cite{Zamir2020CVPR}, and multi-task learning \cite{wang2017transitive,pentina2017multi,misra2016cross}. More specifically, recent attention has focused on using cross-task supervised learning to improve the performance of a single task, such as to improve depth prediction under the supervision of semantic understanding  \cite{guizilini2020semantically,Chen_2019_CVPR}. 
% Since the work of visual attention mechanism \cite{vaswani2017attention} has been published, some recent works also have shown the capability of applying the attention mechanism into multi-task learning  \cite{wang2021domain,jiao2018look,kendall2018multi}.

In this work, we draw inspiration from multi-task learning and single-task failure detection approaches. Our approach utilizes every single task's estimation information, exploiting their latent correlations, and infer a more robust per-task pixel-wise failure prediction.  

\section{Method}
\label{sec:method}
% Introduce the section map:
% 1 Overview of the whole pipeline
% 2 Multi-task element generation
% 3 Attention network 
% 4 Training Procedure
%We propose a general framework to infer task failures. Our approach is inspired by the advantages of multi-task learning and visual attention network. In \ref{subsec:overall}, we first introduce the overall pipeline of our approach. Then in \ref{subsec:data} 
%, we illustrate how we process each task separately, the processed task component will be used as the input to our attention mechanism. Next, we describe our attention network architecture in \ref{subsec:network}. And finally, in \ref{subsec:training}, the training process of our model is covered.
\begin{figure*}[t]
    \centering
    \includegraphics[scale=0.24, trim = 1cm 2cm 1cm 0cm]{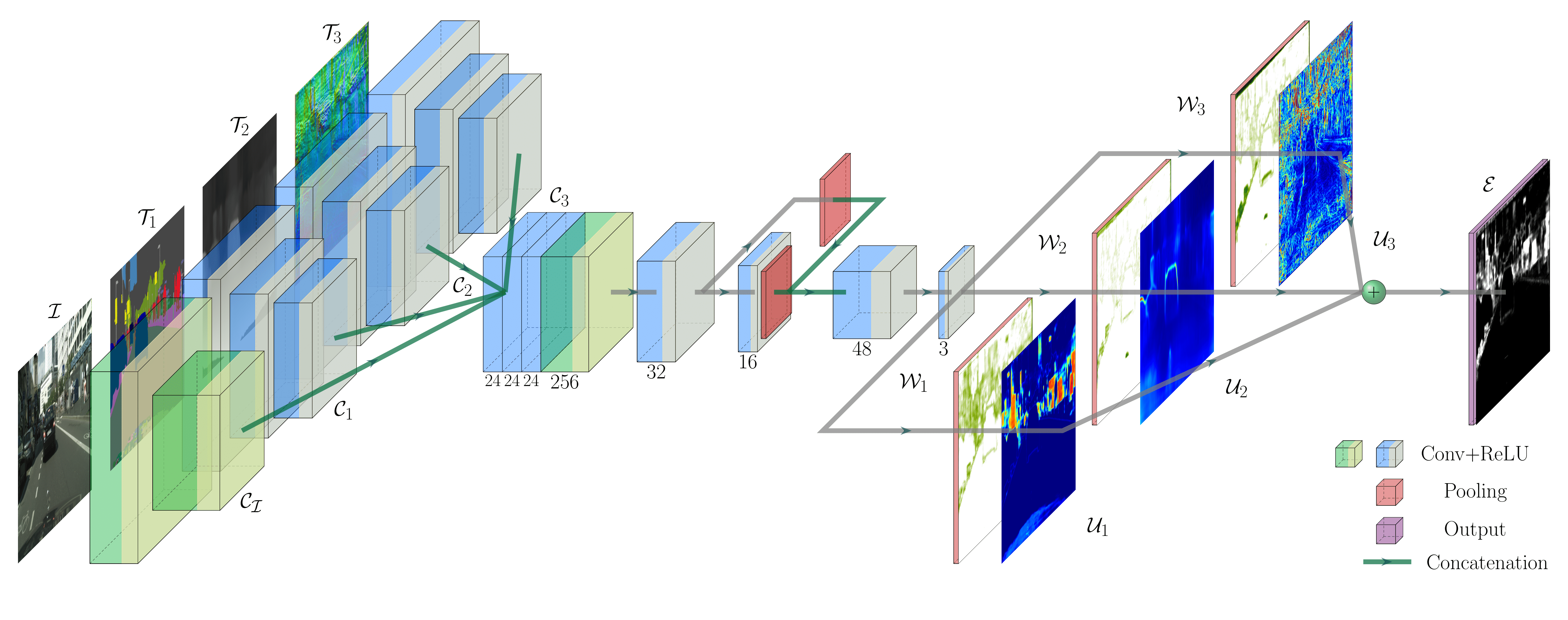}
    \caption{\textbf{Example of the Model Architecture}: This example model uses three different tasks: semantic segmentation, depth, and normal estimation. For the uncertainties $\mathcal{U}_1, \mathcal{U}_2, \mathcal{U}_3$ and prediction errors $\mathcal{E}$, we resize them to $256\times256$. Using the predefined attention patch size $p$, the output attention from the model $\mathcal{W}_1, \mathcal{W}_2, \mathcal{W}_3$ will have the size $(256/p) \times (256/p)$. Then the output would be equally upscaled by a factor of $p$ so that the attention maps' sizes are $256\times256$. Now the computed attention maps have the same size as the uncertainties, then element-wise multiplication can be performed.}
    \label{fig:struct}
    \vspace{-5mm}
\end{figure*}

% Overview of the whole approach, we introduce the idea around Figure 3. We want the reader to understand the whole process: what we produce to feed into the attention network, what is the output from attention network. Details of the network will NOT be covered. 

\subsection{Approach Architecture}
\label{subsec:overall} 
The intuition of our proposed method comes from the fact that tasks can compensate and refine each other's prediction in a multi-task setup\cite{pmlr-v119-standley20a, zamir2018taskonomy}. 
Therefore, it will be easier to identify a certain task prediction failure with the existence of some other tasks. However, unlike a simple high-level task selection case, the proposed approach generates a pixel-wise attention map for each participant task. For clarification, the \textit{Attention} used in this work is a scalar-product and it does not refer to the Transformer Networks \cite{vaswani2017attention} in computer vision community. These maps are used to compute a weighted sum of each task's own failure prediction. Starting with a query image, firstly, predictions of all participant tasks are generated. Secondly, these predictions and the query image are passed into our attention network. The attention network will predict a pixel-wise attention map for each task. Meanwhile, uncertainties\footnote{We use the term uncertainty here to cover all the statistical measurement of prediction reliability, which is often called uncertainty, reliability or dispersion metrics in other works.} of each task prediction are produced. In the final step, the attention maps are applied to these uncertainties to calculate a weighted sum failure prediction for a specific task. The weighted sum result for this task is expected to have better performance, comparing with its uncertainty estimation method. Figure \ref{fig:overall} shows the basic structure of this aforementioned procedure.

% Introduce the first step of the aforementioned pipeline, i.e., how we generate the input elements for the attention network. 
\subsection{Multi-task Element Generation}
\label{subsec:data} 
%\subsubsection*{Datasets}
% \begin{figure}
%     \centering
%     \includegraphics[scale=0.3]{figures/org_uncert_error.png}
%     \includegraphics[scale=0.34]{figures/fog_uncert_all.png}
%     \caption{\textbf{A sample pair of original and foggy scene}. On the left is the original scene, and on the right is the foggy scene. On the top row are the query images; Three perception tasks: semantic segmentation, depth prediction, and normal estimation are on the second row; The corresponding uncertainties of each task's prediction can be found on the third row. On the last row lies the true prediction error of semantic segmentation and depth prediction}
%     \label{fig:uncert_pred}
% \end{figure}

Our approach admits any number of visual perception tasks that provide per-pixel predictions and their per-pixel uncertainties. We rely on open-source state-of-the-art task prediction methods. As for the uncertainty estimation, many options are available. Softmax entropy \cite{lakshminarayanan2016simple}and softmax distance \cite{rottmann2019prediction} can be used for classification tasks, such as semantic segmentation. Sampling-based methods can be used for both classification and regression tasks. Even, the learning-based failure prediction approach for a single task can play the role of uncertainty estimation and be feed into our attention network. Different chosen uncertainty estimation methods will certainly influence the final output, and thus we evaluate their influence in Section~\ref{sec:result}. 
 
% specifically introduce the attention network structure: the neural network structure, size of each layer, I/O, etc.   
\subsection{Attention Network Model}
\label{subsec:network}
We denote the original image as $\mathcal{I}$. The predictions of all $n$ tasks are denoted as  $\mathcal{T}_1, \mathcal{T}_2, \cdots, \mathcal{T}_n$. The uncertainty estimations of them are denoted as $\mathcal{U}_1$,  $\mathcal{U}_2$, $\cdots$, $\mathcal{U}_n$, respectively. 
The architecture of our attention network is shown in Figure \ref{fig:struct}. The model first encodes the original images $\mathcal{I}$ and its task predictions $\left\{\mathcal{T}_i\right\} , i \in \{1, 2 \dots, n \}$. We chose to encode the image $\mathcal{I}$ with the first several layers of  ResNet50 \cite{resnet} into a $256$-channel feature with a $128\times128$ size. On the other hand, the predictions $\left\{\mathcal{T}_i\right\}$ are encoded by part of MobileNetV2 \cite{mobilenet}. Each of them is encoded into a $24$-channel feature map of size $128\times128$. All tasks prediction, $\left\{\mathcal{T}_i\right\}$, share the same encoding structure. The encoded feature maps are denoted as $\mathcal{C_I}$, $\left\{\mathcal{C}_i\right\}$, correspondingly.

After the encoding process, the encoded features are concatenated along the channel dimension. The resulting feature map, $\mathcal{C}_{cat}$ is in the size of $(256+24n)\times128\times128$. $\mathcal{C}_{cat}$ is then forwarded into a neural network with four convolutional layers. In these convolutional layers, the output of the second and third layers will be concatenated together and used as the input to the last layer. In this case, the output layer has one channel for each task. Given a predefined patch size $p$, two pooling layers are added after the second and the third convolutional layers to resize each output channel into $(256/p)\times(256/p)$, which is the resolution of our attention maps.  The higher the required attention resolution, the larger the resized map size of each channel. An extra nearest neighbour rescaling layer is added here to rescale each channel to the size of the uncertainty maps. The rescaled feature maps in each channel are the final attention map generated by our attention network. We denote them as $\mathcal{W}_1$, $\mathcal{W}_2$, $\cdots$, $\mathcal{W}_n$, for each task, respectively. The final failure prediction for a single task is generated by calculating the weighted sum of all tasks' uncertainty estimates, with the attention maps as the weights. The weighted sum failure prediction is denoted as $\mathcal{E}$.

\begin{equation}
    \mathcal{E} = \sum_{i=1}^{n}\mathcal{W}_i \cdot \mathcal{U}_i
    \label{eq:seman_1}
\end{equation}

\subsection{Training Procedure}
\label{subsec:training}
As introduced in the last subsection, the attention maps are predicted by our attention network: 
\begin{equation}
    \left\{\mathcal{W}_i\right\} = f_\mathbf{\theta}(\mathcal{I}, \left\{\mathcal{T}_i\right\})
    \label{eq:net}
\end{equation}
We set a single error approximation loss function to learn the network parameters $\mathbf{\theta}$. When training the network to learn a certain task failure, we compute the pixel-wise prediction error for this task, which is denoted as  $\mathcal{\epsilon}_{\{\cdot\}}$. For example, $\mathcal{\epsilon}_S$ is computed as the cross-entropy for semantic segmentation, and $\mathcal{\epsilon}_D$ is simply calculated as the L2-norm between depth prediction and ground truth depth. The general loss computation is shown as Equation \ref{eq:seman}.

\begin{equation}
    loss_{\{\cdot\}} = \|\sum_{i=1}^n\left(\mathcal{W}_i \cdot \mathcal{U}_i\right) - \mathcal{\epsilon}_{\{\cdot\}}\|
    \label{eq:seman}
\end{equation}

At last, to prevent any task's error from dominating the prediction due to the imbalanced scales, we perform an image-wise normalization process to re-scale all the $\mathcal{C}_{cat}$ in both training and inference stages into range $[0, 1]$.

\begin{table}[t]
\centering 
\begin{tabular}{c|c|c }
\thickhline
{\multirow{2}*{\textbf{Task}}}&\textbf{Prediction}&\textbf{Uncertainty Estimation}\\
&\textbf{Method}&\textbf{Methods}\\
\hline
  & \multirow{4}*{SDC Net \cite{zhu2019improving}}&Softmax Entropy \cite{hendrycks2016baseline}\\Semantic& {}&Softmax Distance \cite{rottmann2019prediction}\\
 Segmentation& {}&Synboost$^*$ \cite{di2021pixel} \\
 & {}&MC Dropout \cite{mcdropout} \\
\hline
 \multirow{2}*{Depth}& \multirow{3}*{Monodepth V2\cite{godard2019digging}}&Bayesian Estimation\cite{neal2012bayesian}\\\multirow{2}*{Estimation}& {} & MC Dropout \cite{mcdropout}
 \\&  & Self Learning$^*$ \cite{Poggi_CVPR_2020} \\
 \hline
Normal & \multirow{2}*{VNL \cite{enforcing}}&\multirow{2}*{Flipping$^*$\cite{godard2017unsupervised}} \\
Estimation&&\\
\hline
Instance & \multirow{2}*{EfficientPS\cite{mohan2020efficientps}}&ROI Softmax  \\
Segmentation&&Uncertainty$^*$ \cite{maskrcnn}\\
\hline
\thickhline % Bottom horizontal line
\end{tabular}
\caption{The selected task prediction methods and the uncertainty estimation methods for all different tasks in the experiments. $^*$ indicates the method used by default in the experiments unless otherwise mentioned.} 
\label{tab:methods} 
\vspace{-5mm}
\end{table}

\section{Experiments}
\label{sec:result}

\subsection{Experimental Setup}
\label{subsec:config}

\subsubsection{Model Implementation} % \textbf{Model Implementations}
\label{subsubsec:implement}
Inspired by task networks graph in \textit{taskonomy} \cite{zamir2018taskonomy}, we decided to choose tasks from three visual tasks in total in our different experiments: semantic segmentation, depth estimation, and normal estimation. We add later in the experiments a fourth task, instance segmentation. For each task, we implemented publicly available task prediction methods and uncertainty estimation methods. All evaluated methods are shown in Table \ref{tab:methods}.

\subsubsection{Dataset}
\label{subsubsec:dataset}
Training was performed on Cityscapes training dataset \cite{Cordts2016Cityscapes}, including 2975 driving scenes images with fine semantic annotations and disparity ground truth. We produced our dataset by applying the methods shown in the Table \ref{tab:methods}. The training set is then composed by set in form of $\{\mathcal{I}, \mathcal{T_S}, \mathcal{T_D}, \mathcal{T_N}, \mathcal{U_S}, \mathcal{U_D}, \mathcal{U_N}, \mathcal{\epsilon}_{\{\cdot\}}\}$, where $S,D,N$ denote semantic segmentation, depth estimation, and normal estimation, tasks, respectively, and $\mathcal{\epsilon}_{\{\cdot\}}$ correspond to the prediction error of the chosen task to predict its failure.

To test our model's performance, pre-processing of the various test datasets is also required. Here we performed the same pipeline as mentioned in the subsection \ref{subsubsec:dataset} for Cityscapes validation set \cite{Cordts2016Cityscapes}, Foggy Cityscapes validation set \cite{foggyCityscapes}, Wilddash \cite{Zendel_2018_ECCV} and Dark Zurich dataset \cite{darkzurich}. Wildash provides a dataset and benchmark for challenging driving scenarios under real-world conditions, it contains scenarios from very diverse environments, locations, and weather conditions. Dark Zurich is a dataset designed for semantic uncertainty-aware model evaluations. It contains driving scenes images captured at night time, twilight and day time. Here we only use the night time images for our evaluation. The purpose of testing on these extra two datasets is to validate the model robustness when dealing with the \emph{challenging unseen scenarios}.

\subsubsection{Metrics}
We choose the zero-mean normalized cross-correlation (ZNCC) in our experiments as a measurement of how close the predicted failure is to the ground-truth failure.
The ZNCC of the estimation $\mathcal{E}$ and the ground truth error $\mathcal{\epsilon}$ is defined as:
\begin{equation}
    \textrm{ZNCC} = \frac{\sum_{(u, v)}(\mathcal{E}(u,v) - \mu_{\mathcal{E}})(\mathcal{\epsilon}(u,v)- \mu_{\mathcal{\epsilon}})} {\sqrt{\sum_{(u, v)}(\mathcal{E}(u,v- \mu_{\mathcal{E}})^2}\sqrt{\sum_{(u, v)}(\mathcal{\epsilon}(u,v))- \mu_{\mathcal{\epsilon}})^2}}
\end{equation}
\noindent where $\mu_{\mathcal{E}}$ and $\mu_{\mathcal{\epsilon}}$ denote the mean values of $\mathcal{E}, \mathcal{\epsilon}$, respectively, and $(u,v)$ are the pixel locations.

ZNCC is invariant of affine pixel value changes \cite{zncc}. Therefore, it is not be influenced by the normalization operation during the prediction and evaluation procedure, and it is also less sensitive to resizing operations than normalized cross-correlation (NCC).
This metric is seamlessly applicable to classification and regression tasks.

In addition, for the classification task of semantic segmentation, previous works on failure detection have used several metrics for evaluation \cite{di2021pixel,marufur2021fsnet,haldimann2019not}. Thus, we also report:
\begin{itemize}
    \item AUPR-Error: the area under the Precision-Recall (AUPR) curve, which regard incorrect prediction as the positive class.
    \item AUPR-Success: it computes AUPR as well, whereas treats correct prediction as the positive class.
    \item FPR95: the false positive rate at 95\% true positive rate.
\end{itemize}
As for regression tasks, such as depth estimation, we are not aware of any previous work focusing on evaluating the failure prediction model. 

\subsection{Comparisons}

Our framework is built on a multi-task setup, and uses different single task's uncertainty estimation methods as the input. Thus, most of the our evaluations focus on relative comparison between our model's output and the input it uses, or among the models with different configurations. We set up several experiments to evaluate different components or factors and be able to answer a series of questions.% These questions, together with experiments and analyse towards answering them will be discussed one by one.

% \begin{itemize}
%     \item First and foremost, are multiple tasks beneficial for single task failure detection?
%     \item Is the multiple task setup useful for failure detection in both classification and regression tasks?
%     \item At what resolution should the attention maps be computed to capture the relevant information for failure detection?
%     \item How dependant is our failure detection on the uncertainty estimate input?
%     \item How is the failure detection performing when adding one more task?
%     \item How is our failure detection generalizing to scenarios with larger distribution mismatch?
% \end{itemize}

% We now conduct the experiments and analyse the results towards answering these questions.
%The evaluation result of each group is displayed and discussed in section \ref{subsec:result}.  

\medskip

\textbf{%Multiple Tasks.
Are multiple tasks beneficial for single task failure detection?}
In these experiments we evaluate two aspects. The first one is the influence of increasing the number of tasks as inputs to our failure detection. And the second one, we evaluate our failure detection on different main tasks: a classification task (semantic segmentation) and a regression task (depth estimation).
\begin{table}[t]
\centering
\setlength{\tabcolsep}{5pt}
%\scalebox{0.965}{
\begin{tabular}{c c c|c c c c c c}
\thickhline
 \multicolumn{3}{c|}{\textbf{Task Entries}} & \multicolumn{4}{c}{\textbf{Semantic}} &\textbf{Depth}\\ \hline
$S$&$D$&$N$&ZNCC$\uparrow$ & AP-Err$\uparrow$  & AP-Suc$\uparrow$&FPR95$\downarrow$ &ZNCC$\uparrow$ \\ % Column names row
\hline  

\checkmark&\checkmark&\checkmark& \textbf{0.649} & \textbf{0.590} &0.987&0.280 &\textbf{0.646} \\ % Content row 1
\checkmark&\checkmark&& 0.609 & 0.545 &\textbf{0.990}&\textbf{0.278} & 0.489 \\ % Content row 2

\checkmark&&&0.494& 0.413&0.978&0.570 & - \\ % Content row 2
&\checkmark&& - &-& &-&0.483\\ % Content row 2
\thickhline
\end{tabular}
%}
\caption{Multiple Tasks Experiments: Comparison among multiple different task entries for both semantic segmentation's and depth estimation's failure prediction. The check-mark represents the task at the corresponding place is included in the model, both during training and evaluation process.} 
\label{tab:multi-task} 
\vspace{-5mm}
\end{table}

We start in our network structure by only having the input of a single task (the task for which the failure is being detected), see last 2 rows of Table~\ref{tab:multi-task}, on the Cityscapes original validation set. This is equivalent to learning failures from single task knowledge and thus can be compared to use the uncertainty input as a proxy to the failure. 

Then, we continue by adding a second task (depth or semantics, as appropriate), and a third one (normal estimation), see first row in Table~\ref{tab:multi-task}.

From this experiment we have evidence that, indeed, a multi-task setup improves failure detection of one task, and, this is true for both semantic segmentation and depth estimation. This is a confirmation of our hypothesis that our framework leverages the latent correlations among tasks to improve the introspection capabilities of the every single one of them. It is also in line of the findings of \cite{zamir2018taskonomy} where is shown that the normal estimation task is more correlated to the depth estimation than to the semantic segmentation one, as seeing in the higher increase in performance of the depth estimation failure detection compared to that improvement the semantic segmentation failure detection.

\medskip

\textbf{%Attention Maps Resolution. 
At what resolution should the attention maps be computed to capture the relevant information for failure detection?}
In our network architecture, we have the choice to generate the attention maps at different resolutions by modifying the last pooling layers. Something that could be beneficial for differentiating between full image task failure or per pixel or region task failure detection.  

\begin{table}[h]
\centering
\begin{tabular}{c | c |c c  | c c}
\thickhline
\multirow{2}{*}{\textbf{Method}}&\multirow{2}{*}{\textbf{Patch}}&\multicolumn{2}{c}{Original}&\multicolumn{2}{c}{Foggy}\\
\cline{3-6}
&&ZNCC$\uparrow$ &AP-Err$\uparrow$& ZNCC$\uparrow$ &AP-Err$\uparrow$\\  % Column names row
\hline
\multirow{8}{*}{Ours}&1 &  \textbf{0.649} & \textbf{0.590} & \textbf{0.560} & \textbf{0.518} \\ % Content row 1
&2 & 0.601 & 0.520 & 0.556 & 0.495 \\ % Content row 1
&4 & 0.570 & 0.487 & 0.530 & 0.475  \\ % Content row 2
&8 & 0.529 & 0.439 & 0.530 & 0.470 \\ % Content row 3
&16 & 0.489 & 0.400 & 0.516& 0.466  \\ % Content row 4
&32 &0.455 & 0.368 & 0.500 & 0.450\\ % Content row 5
&64 & 0.440& 0.356 & 0.496 & 0.448\\ % Content row 5
&128 & 0.437 & 0.357 & 0.501 & 0.454\\ % Content row 5
\hline
SynBoost& -&0.450 & 0.387 & 0.506& 0.480\\
\thickhline % Bottom horizontal line
\end{tabular}% A label for referencing this table elsewhere, references are used in text as \ref{label}
\caption{Variable Attention Map Resolution: Comparison among multiple different patch sizes for semantic segmentation's failure prediction.}
\label{tab:multi-patch_s}
\vspace{-5mm}
\end{table}

\begin{table}[h]
\centering
\begin{tabular}{c | c |c | c }
\thickhline
\multirow{2}{*}{\textbf{Method}}&\multirow{2}{*}{\textbf{Patch}}&\multicolumn{1}{c}{Original}&\multicolumn{1}{c}{Foggy}\\
\cline{3-4}
&&ZNCC$\uparrow$ & ZNCC$\uparrow$ \\  % Column names row
\hline
\multirow{8}{*}{Ours}&1 & \textbf{0.646}  & 0.570 \\ % Content row 1
&2 &0.638  & \textbf{0.573} \\ % Content row 1
&4 & 0.627&  0.560 \\ % Content row 2
&8 & 0.611 & 0.537  \\ % Content row 3
&16 & 0.569 & 0.529  \\ % Content row 4
&32 &0.455 &  0.445\\ % Content row 5
&64 & 0.317 & 0.341 \\ % Content row 5
&128 & 0.279 & 0.282 \\ % Content row 5
\hline
Self Learning&- & 0.248 &0.255\\
\thickhline
\end{tabular}
\caption{Variable Attention Map Resolution: Comparison among multiple different patch size for depth estimation's failure prediction.}
\label{tab:multi-patch_d}
\vspace{-4mm}
\end{table}

We modify the patch size of our model from 1 to 128, while the generated attention map is up-scaled to size $256\times 256$. From the evaluation results in Tables~\ref{tab:multi-patch_s} and \ref{tab:multi-patch_d}, we can conclude that with higher resolution the model is able to predict failures more accurately, for it has higher AP-Err, and higher ZNCC for both task failures. Here we included as well the Foggy dataset to check the performance in more challenging scenarios.

\medskip

\textbf{%Uncertainty Input. 
How dependant is our failure detection on the uncertainty estimate input?}
The uncertainty of each task is an important component of our failure detection framework. Thus, this experiment investigates whether our conclusions have been biased to the uncertainty input used. For the semantic segmentation task we evaluate three more uncertainty inputs, and for the depth estimation another two (see Table~\ref{tab:methods}). For each specific uncertainty method, we select two of our models with different patch sizes (1 and 16).

\begin{table}[h]
\centering 
\begin{tabular}{c | c |c c  | c c  }
\thickhline\multirow{2}{*}{\textbf{Method}}&\multirow{2}{*}{\textbf{Patch}}&\multicolumn{2}{c}{Cityscapes Original}&\multicolumn{2}{c}{Cityscapes Foggy}\\
\cline{3-6}
&&ZNCC$\uparrow$ &AP-Err$\uparrow$& ZNCC$\uparrow$ &AP-Err$\uparrow$\\  % Column names row
\hline
Ours with &1 &  \textbf{0.649} & \textbf{0.590}  & \textbf{0.560} & \textbf{0.518}   \\ % Content row 1
Synboost&16 & 0.489 & 0.400 & 0.516& 0.466  \\ % Content row 1
\hline
SynBoost& -&0.450 & 0.387  & 0.506& 0.480  \\
\thickhline
Ours with &1 & \textbf{0.681} & \textbf{0.618}  & \textbf{0.628} & \textbf{0.565}  \\ % Content row 1
Soft. Ent.&16 & 0.568 & 0.447  & 0.593 & 0.497  \\ % Content row 1
\hline 
Soft. Ent.&- & 0.572 & 0.444& 0.619 & 0.520\\
\thickhline
Ours with &1 & \textbf{0.576} & \textbf{0.506}  & \textbf{0.430} & \textbf{0.408}  \\ % Content row 1
MC Dropout&16 & 0.358& 0.274 & 0.327& 0.307  \\ % Content row 1
\hline
MC Dropout& -&0.249 & 0.218  & 0.163& 0.236 \\
\thickhline
Ours with &1 & \textbf{0.668} & \textbf{0.593} & \textbf{0.600} & \textbf{0.532} \\ % Content row 1
Soft. Dis&16 & 0.540 & 0.409 &0.574  & 0.479  \\ % Content row 1

\hline 
Soft. Dis.& -&0.527 & 0.408  & 0.569 & 0.489  \\
\thickhline
\end{tabular}
\caption{Effect of Changing the Semantic Uncertainty Input: Our models vs. selected uncertainty inputs for semantic estimation's failure prediction.}
\label{tab:multi-uncert} 
\vspace{-4mm}
\end{table}

\begin{table}[h]
\centering 
\begin{tabular}{c | c |c| c  }
\thickhline
\multirow{2}{*}{\textbf{Method}}&\multirow{2}{*}{\textbf{Patch}}&Original&Foggy\\
\cline{3-4}
&&ZNCC$\uparrow$ &ZNCC$\uparrow$\\  % Column names row
\hline
Ours with &1 &\textbf{0.646}  & \textbf{0.570}\\ % Content row 1
Self Learning&16 & 0.569 & 0.529  \\ % Content row 1
\hline 
Self Learning&- & 0.248 & 0.255\\
\thickhline
Ours with &1 & \textbf{0.757} & \textbf{0.645} \\ % Content row 1
Bayesian&16 & 0.684 & 0.584 \\ % Content row 1
\hline
Bayesian&-&0.091 & 0.074 \\
\thickhline
Ours with &1 & \textbf{0.648} & \textbf{0.516} \\ % Content row 1
MC Dropout&16 & 0.575 & 0.446  \\ % Content row 1
\hline
MC Dropout&- & 0.076 & 0.075\\
\thickhline% Bottom horizontal line
\end{tabular}
\caption{Effect of Changing the Depth Uncertainty Inputs: Our models vs. selected uncertainty methods for depth estimation's failure prediction. %ZNCC is used as evaluation metric.
}
\label{tab:multi-uncert-d}
\vspace{-4mm}
\end{table}

The results of this investigation can be seen in Tables~\ref{tab:multi-uncert} and \ref{tab:multi-uncert-d}. Our framework is consistently outperforming the uncertainty input for both tasks failure detections, within both original and foggy image set from Cityscapes.
\begin{figure*}
    \centering
    \includegraphics[scale=0.473]{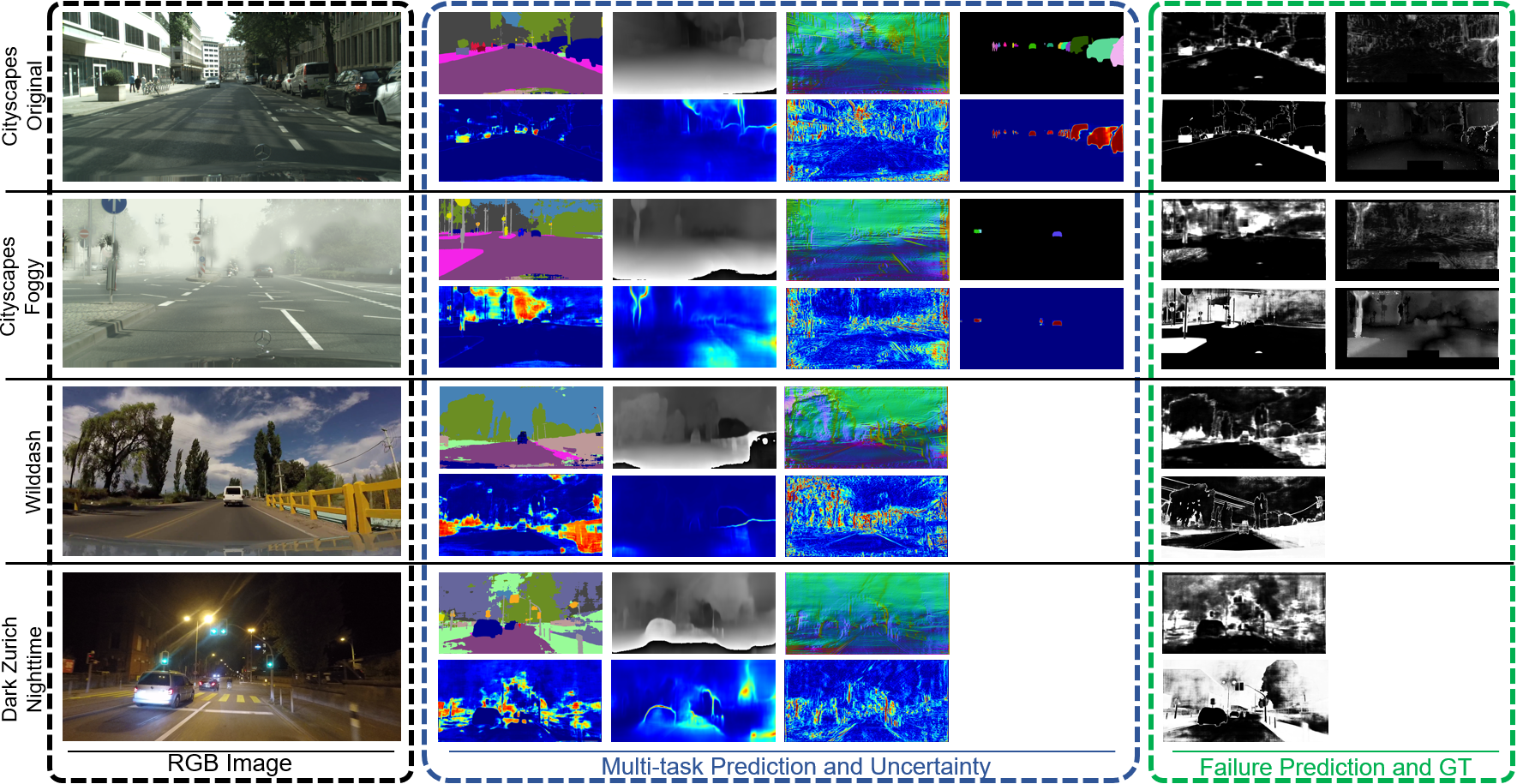}
    \caption{\textbf{Qualitative Examples}. For an input image in the left block, the middle block contains the various visual tasks ($1^{st}$ row) and the corresponding uncertainties ($2^{nd}$ row), tasks are semantic, depth, normal, instance, from left to right. In the right block our failure predictions ($1^{st}$ row) and the ground truth ($2^{nd}$ row) are presented, semantic on the left and depth on the right. For the purpose of clear visualization, the ego-vehicle part in both Cityscapes examples are filtered out.}
    \label{fig:qualitative_result}
    \vspace{-5mm}
\end{figure*}

\medskip

\textbf{%One More Task. 
How is the failure detection performing when adding one more task?}
Finally, the question comes to our failure detection based on a multi-task setup is fixed to the already chosen tasks in the previous experiments. For that reason, we add the extra task of instance segmentation ($IS$), with its corresponding uncertainty estimate, see last row of Table~\ref{tab:methods}.

\begin{table}[h]
\centering
\begin{tabular}{c|c c c c|c c c }
\thickhline
 \multirow{2}{*}{\textbf{Patch}}&\multicolumn{4}{c|}{\textbf{Task Entries}} &\multicolumn{2}{c}{\textbf{Semantic}} &\textbf{Depth}\\ 
&$S$&$D$&$N$&$IS$&ZNCC$\uparrow$ & AP-Err$\uparrow$ &ZNCC$\uparrow$  \\ % Column names row

\hline
 \multirow{2}{*}{1}&\checkmark&\checkmark&\checkmark&\checkmark & 0.641 & 0.585  &\textbf{0.655} \\ 
&\checkmark&\checkmark&\checkmark& & \textbf{0.649} & \textbf{0.590} &0.646\\
\hline
 \multirow{2}{*}{16}&\checkmark&\checkmark&\checkmark&\checkmark & \textbf{0.493} & \textbf{0.403}  &\textbf{0.581} \\ 
&\checkmark&\checkmark&\checkmark& & 0.489 & 0.400 &0.529 \\ % \thickhline
\thickhline
\end{tabular}
\caption{Adding an Extra Task: Comparison among multiple different task entries for both semantic segmentation's and depth estimation's failure prediction.} 
\label{tab:extra-task}
\vspace{-4mm}
\end{table}

The results of this experiment, shown in Table~\ref{tab:extra-task}, indicate that the addition of 
an extra task continues to be beneficial for the different tasks failure detections. However, we observe that the contribution is higher for the depth estimation failure detection than for the semantic segmentation. We believe this is due to less extra information provided by the instance segmentation with respect to the semantic segmentation task. While differentiating among different instances of the same class is highly informative for the depth estimation task.

\medskip

\textbf{%More Challenging Scenarios. 
How is our failure detection generalizing to scenarios with larger distribution mismatch?}
Here, we use our default models trained with the Cityscapes dataset, and deployed them on Wilddash and Dark Zurich (night) datasets, for the tasks of failure detection of the semantic segmentation. Results can be seen in Table~\ref{tab:multi-set}. Additionally, we include the evaluation of different uncertainty inputs as they are quite dependent on the distribution mismatch between the test and training set. We can conclude that the improvements brought by our framework generalize to more complex scenarios, invariant to the uncertainty estimate input.

\begin{table}[h]
\centering 
\begin{tabular}{c  |c c | c c  }
\thickhline\multirow{2}{*}{\textbf{Method}}&\multicolumn{2}{c}{Wilddash}&\multicolumn{2}{c}{Dark Zurich}\\
\cline{2-5}
&ZNCC$\uparrow$ &AP-Err$\uparrow$& ZNCC$\uparrow$ &AP-Err$\uparrow$\\  % Column names row
\hline
Ours with Synboost  & \textbf{0.412} & \textbf{0.595} & \textbf{0.238} &  \textbf{0.775}  \\ % Content row 1
\hline
SynBoost& 0.323 & 0.584 & 0.199 & 0.726  \\
\thickhline
Ours with Soft. Ent. & \textbf{0.520} & \textbf{0.630} & 0.544 & \textbf{0.867}  \\ % Content row 1
\hline 
Soft. Ent. & 0.508 & 0.625 & \textbf{0.578}& 0.830 \\
\thickhline
Ours with MC Dropout & \textbf{0.321} &\textbf{ 0.537} &\textbf{0.101} & \textbf{0.734}  \\ % Content row 1
\hline
MC Dropout&0.139& 0.428 & -0.128 & 0.678 \\
\thickhline
Ours with Soft. Dis & \textbf{0.511} & \textbf{0.628} & 0.497 & \textbf{0.861}\\ % Content row 1

\hline 
Soft. Dis.&0.478 & 0.619& \textbf{0.502} & 0.797 \\
\thickhline
\end{tabular}
\caption{Generalization: Our models vs. selected uncertainty inputs for semantic segmentation's failure prediction on two other datasets: Wilddash and Dark Zurich.}
\label{tab:multi-set} 
\vspace{-5mm}
\end{table}

Some qualitative results for the different datasets are visualized in figure \ref{fig:qualitative_result}. 

\section{Conclusion}
\label{sec:conclusion}
We propose a framework to detect visual task prediction failures. We leverage the information from multiple visual tasks simultaneously being deployed, and build a learning-based attention neural network to perform a weighted sum of task uncertainties to approximate the task prediction failure. Our approach is more accurate in detecting semantic and depth prediction errors, compared with various uncertainty estimation methods. Additionally, our thorough experimental evaluation also proves its ability to further improve the performance by increasing the attention map resolution, as well as by including in extra correlated visual tasks. 
Finally, we observe that the multi-task setup allows for better generalization to environments with a larger distribution mismatch to that of the training set.
We believe our framework shows the potential capability to be applied on various autonomous mobile robots with multi-task visual modules, such as safety in autonomous driving. 

\bibliography{example}

\end{document}